\documentclass[10pt,twocolumn,letterpaper]{article}

\usepackage{cvpr}              

\usepackage{graphicx}
\usepackage{amsmath}
\usepackage{amssymb}
\usepackage{wasysym}
\usepackage{booktabs}
\usepackage{multirow}
\usepackage{bm}
\usepackage{caption}
\usepackage{subcaption}
\usepackage[accsupp]{axessibility}

\usepackage[pagebackref,breaklinks,colorlinks]{hyperref}

\usepackage[capitalize]{cleveref}
\crefname{section}{Sec.}{Secs.}
\Crefname{section}{Section}{Sections}
\Crefname{table}{Table}{Tables}
\crefname{table}{Tab.}{Tabs.}

\def\cvprPaperID{3205} 
\def\confName{CVPR}
\def\confYear{2023}

\begin{document}

\title{Learning Instance-Level Representation for Large-Scale Multi-Modal Pretraining in E-commerce}
\author{Yang Jin\textsuperscript{1,2}, Yongzhi Li\textsuperscript{2}, Zehuan Yuan\textsuperscript{2}, Yadong Mu\textsuperscript{1\thanks{Corresponding Author.}}\\
\textsuperscript{1}Peking University \quad
\textsuperscript{2}ByteDance Inc. \\
\tt\small jiny@stu.pku.edu.cn, \tt\small liyongzhi.ailab@bytedance.com, \\
\tt\small yuanzehuan@bytedance.com, \tt\small myd@pku.edu.cn
}

\maketitle

\begin{abstract}
This paper aims to establish a generic multi-modal foundation model that has the scalable capability to massive downstream applications in E-commerce. Recently, large-scale vision-language pretraining approaches have achieved remarkable advances in the general domain. However, due to the significant differences between natural and product images, directly applying these frameworks for modeling image-level representations to E-commerce will be inevitably sub-optimal. To this end, we propose an instance-centric multi-modal pretraining paradigm called ECLIP in this work. In detail, we craft a decoder architecture that introduces a set of learnable instance queries to explicitly aggregate instance-level semantics. Moreover, to enable the model to focus on the desired product instance without reliance on expensive manual annotations, two specially configured pretext tasks are further proposed. Pretrained on the 100 million E-commerce-related data, ECLIP successfully extracts more generic, semantic-rich, and robust representations. Extensive experimental results show that, without further fine-tuning, ECLIP surpasses existing methods by a large margin on a broad range of downstream tasks, demonstrating the strong transferability to real-world E-commerce applications.
\end{abstract}

\section{Introduction}
\label{sec:intro}

Nowadays, the flourishing growth of E-commerce has brought great convenience to people's daily life. And a wide range of product-based application tasks has subsequently emerged, such as item classification~\cite{xu2019open, pawlowski2022machine}, product retrieval~\cite{zhan2021product1m,gao2020fashionbert}, commodity recommendation~\cite{rahayu2017systematic,wong2021improving}, and so on. Compared to developing individual task-specific models, building a general-purpose foundation model that works for massive E-commercial applications simultaneously can enhance applicability and reduce training costs.

\begin{figure}[t]
  \centering
  \includegraphics[width=0.9\linewidth]{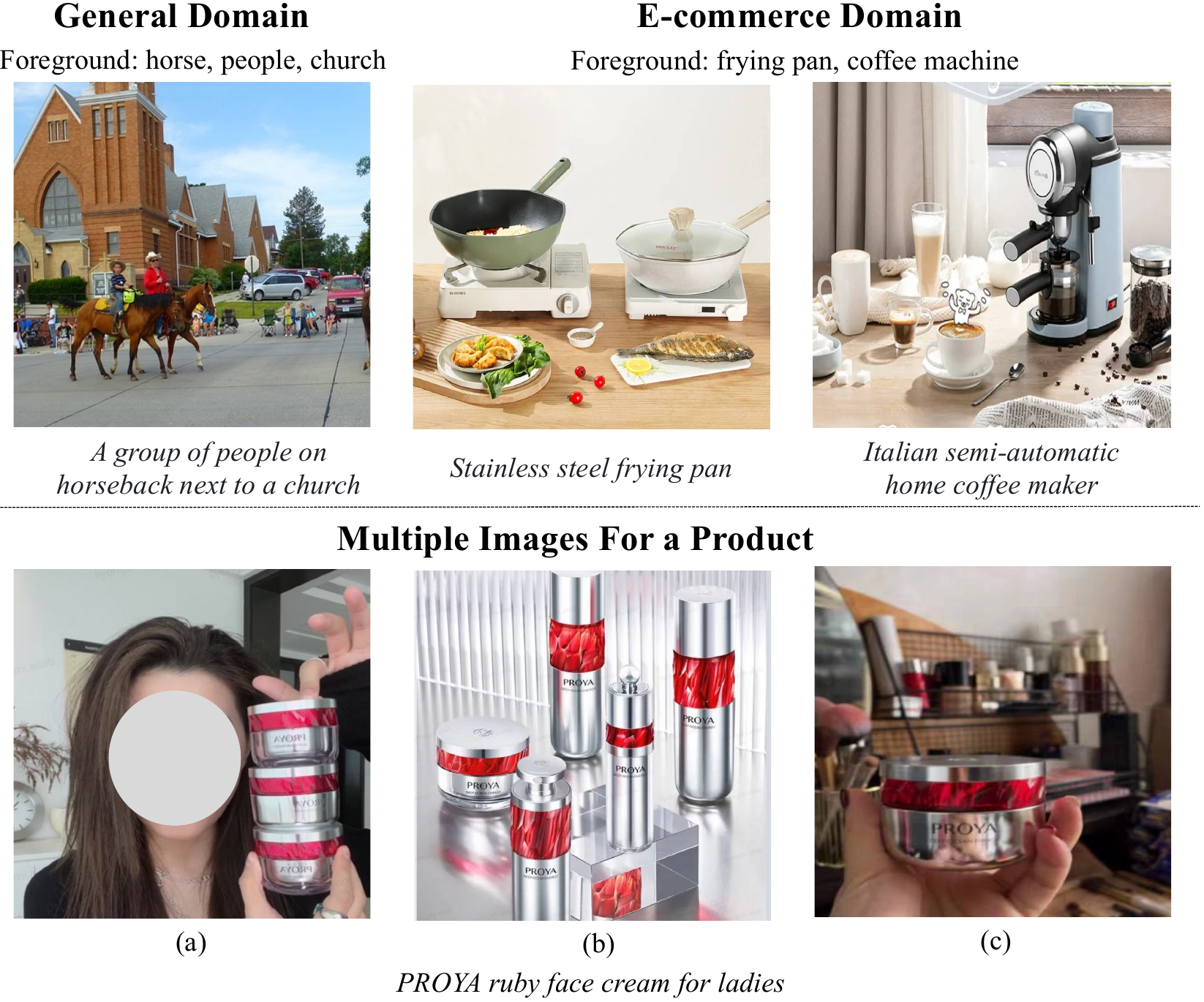}
  \vspace{-0.5em}
  \caption{\small Domain difference between natural and product images. For natural images, it is the frequent case that most pixels are semantically correlated to the textual sentence. However, in E-commerce, such correlation is much more sparse (\eg, ``frying pan" or "coffee machine" only occupy small portions of the entire images). Moreover, images for a product are often provided in a group from multiple sources such as (a)  advertisement videos, (b) product pages, (c) customer comments (see the bottom examples).}
  \label{fig:fig1}
  \vspace{-1.6em}
\end{figure}

Recent developments in vision-language pretraining (VLP)~\cite{lu2019vilbert,jia2021scaling,radford2021learning,li2021align, yao2021filip,zeng2021multi} have demonstrated remarkable advances in diverse VL downstream tasks. Profiting from large-scale image-text pairs, these methods are able to learn generic multimodal representations that are reused across various tasks. In E-commerce scenario, the related data naturally contains cross-modal information to describe a corresponding product. Motivated by the tremendous success achieved by VL modeling, several approaches~\cite{zhan2021product1m,zhu2021knowledge,dong2022m5product,yu2022commercemm} have made attempts at designing a commerce-specific multimodal representation learning paradigm. They imitate the existing VLP methods (e.g., CLIP~\cite{radford2021learning}, VilBERT~\cite{lu2019vilbert}) to learn the \textit{image-level} representations of the product via pretraining on abundant commerce image-text pairs. 

Though promising results have been achieved, directly applying these VLP methods in the general domain to E-commerce still suffers from inherent deficiencies. The properties of natural and product images appear to be dramatically different. Given a natural image-text pair, almost every pixel in the natural image is mentioned by the corresponding textual description. In contrast,  as shown in Figure~\ref{fig:fig1}, in a real E-commerce scenario, the images are mostly product-oriented. Only very few instances are related to the product description. Simply treating the whole image as a monolithic entity to perform cross-modal alignment with text will inevitably confound the foreground and noisy background. Hence, to establish a foundation model that generalizes well to diverse E-commerce applications, it is of great significance to learn the product-related instance-level representation. With this goal in mind, a crucial challenge needs to be addressed: How can we enable the model to focus on the product instance in the presence of background interference? 

A straightforward way to tackle this problem would be to resort to object-level human annotations, but it is laborious and infeasible to scale on larger data from the Internet. In this work, we strive to derive the capability of grounding product instances from uncurated data. Our motivation is built on the natural characteristics of E-commerce data itself. As illustrated in Figure~\ref{fig:fig1}, a product usually has multiple image samples from different sources (\eg, merchant, customer comments, attached advertisement videos, etc.). Although the appearance of these samples may be diverse due to the changes of camera view or scenes, they all include the identical product entity. This fact strongly spurs us to pursue an instance-centric multi-modal learning paradigm by leveraging such explicit correlation. 

The proposed pretraining framework, dubbed as \textbf{ECLIP} (E for ``E-commerce"), employs two separate encoders to embed the images and texts of products. Our key idea is to develop a decoder architecture built upon the above-mentioned encoders, which aims to aggregate the instance-centric product representations without additional hand-crafted annotation. Inspired by~\cite{locatello2020object,carion2020end,xu2022groupvit}, the decoder introduces a set of learnable tokens that we refer to as \textit{instance query}. At each decoder block, these instance queries are updated via interacting with the encoded visual features. Through the stack of multiple blocks, they will gradually probe the potential product instance from the entire image. Moreover, each instance query is conditioned on a concrete text or image called \textit{multi-modal prompt}. Such a design renders it dedicated to a particular instance type indicated by the content of its associated prompt. Therefore, by specifying the content of multi-modal prompt, the decoder can adaptively discover the corresponding instance. During pretraining, there is only one positive prompt for a given sample. The rest are negative ones sampled from other products. 

To effectively optimize the generated instance representations, we newly craft two pretext tasks: inter-product and intra-product multi-modal learning. The first one is in charge of pulling the representations of the identical product closer to each other and pushing away the unmatched ones. It is noteworthy that the appearance of the positive image samples varies a lot except for the presented product. Bringing their representations closer than negative pairs in the feature space will implicitly encourage the instance query to focus on the visual region that corresponds to the desired product. The second one aims to ensure that only positive queries can aggregate the semantics of the foreground instance, rather than negative ones. Coupling these two novel pretext tasks together, we find that the whole framework is capable of learning a generic product representation. Our core contributions can be summarized as follows: 

(1) We propose ECLIP, an effective and simple multi-modal representation learning paradigm in the E-commerce scenario. Going beyond regular global representations, it can successfully obtain instance-centric product representations via a decoder architecture.

(2) By fully exploiting the natural characteristics of E-commerce data and the proposed pretext tasks, ECLIP obtains the fine-grained alignment capability to ground the desired product instance (see Figure~\ref{fig:attn}) without reliance on any manual annotation. 

(3) Pre-trained on large-scale product data, the resulting foundation model can seamlessly generalize to downstream E-commerce applications. Comprehensive experimental results further demonstrate the superiority of ECLIP: without any fine-tuning, it achieves substantial improvements over the existing state-of-the-art methods on diverse real-world E-commerce tasks.

\section{Related Work}
\label{sec:relatd}

\noindent \textbf{Vision-Language Representation Learning.} 
In recent years, vision-language pretraining (VLP) has attracted the attention of numerous researchers and has been widely explored~\cite{gan2022vision}, which aims to learn from tremendous image-text paired data to obtain knowledge that can be generalized to downstream tasks. Some pioneer works (e.g. LXMERT~\cite{tan2019lxmert}, UNITER~\cite{chen2020uniter}, VinVL~\cite{zhang2021vinvl}) rely on pretrained object detection modules such as Faster-RCNN~\cite{ren2015faster} to extract visual representations. Later efforts such as ViLT~\cite{kim2021vilt} and VLMo~\cite{wang2021vlmo} unify the vision and language transformers, and train a multimodal transformer from scratch. Then, CLIP~\cite{radford2021learning} and ALIGN~\cite{jia2021scaling} demonstrate that dual-encoder models pretrained with contrastive objectives on noisy image-text pairs can learn strong image and text representations for crossmodal alignment tasks and zero-shot image classification. While ALBEF~\cite{li2021align} additionally trains a fusion-encoder to jointly learn the multi-modal representations. GLIP~\cite{Li_2022_CVPR} unifies object detection and phrase grounding for pretraining and surpasses many baselines in the detection field. Another line of researches~\cite{wang2021simvlm,wang2022unifying,piergiovanni2022answer,yu2022coca} develop encoder-decoder models that are trained using generative losses and show strong generation performances in vision-language benchmarks, while the visual encoder still performs competitively on image classification. But most aforementioned VLP methods devote themselves to the coarse correlation between the text and the entire image and ignore the instance-level information, which is critical in an e-commerce scenario (as shown in Figure~\ref{fig:fig1}). 

\noindent \textbf{MultiModal Pre-training for E-commerce.}
Early works like FashionBERT~\cite{gao2020fashionbert}, Kaleido-BERT~\cite{zhuge2021kaleido} leverage a transformer-based model and tailored masking strategy to perform pretraining to generate more fine-grained features for cloth retrieval. Then CAPTURE~\cite{zhan2021product1m} generates discriminative instance features via masked multi-modal learning as well as cross-modal contrastive pretraining achieve surprising performance in the instance-level product retrieval task.
K3M~\cite{zhu2021knowledge} further introduces the knowledge modality in multi-modal pretraining to correct the noise and supplement the missing of image and text modalities. 
SCALE~\cite{dong2022m5product} proposes a self-harmonized contrastive learning framework that can integrate six different modalities into a unified model. Recent CommerceMM in~\cite{yu2022commercemm} design a contrastive and MLM-based pretraining paradigm on 14 different tasks. However, all the existing methods only consider the global alignment between images and text, without exploring the special characteristic contained in the e-commercial data for learning instance-centric representation.

\begin{figure*}[t]
  \centering
  \includegraphics[width=0.93\linewidth]{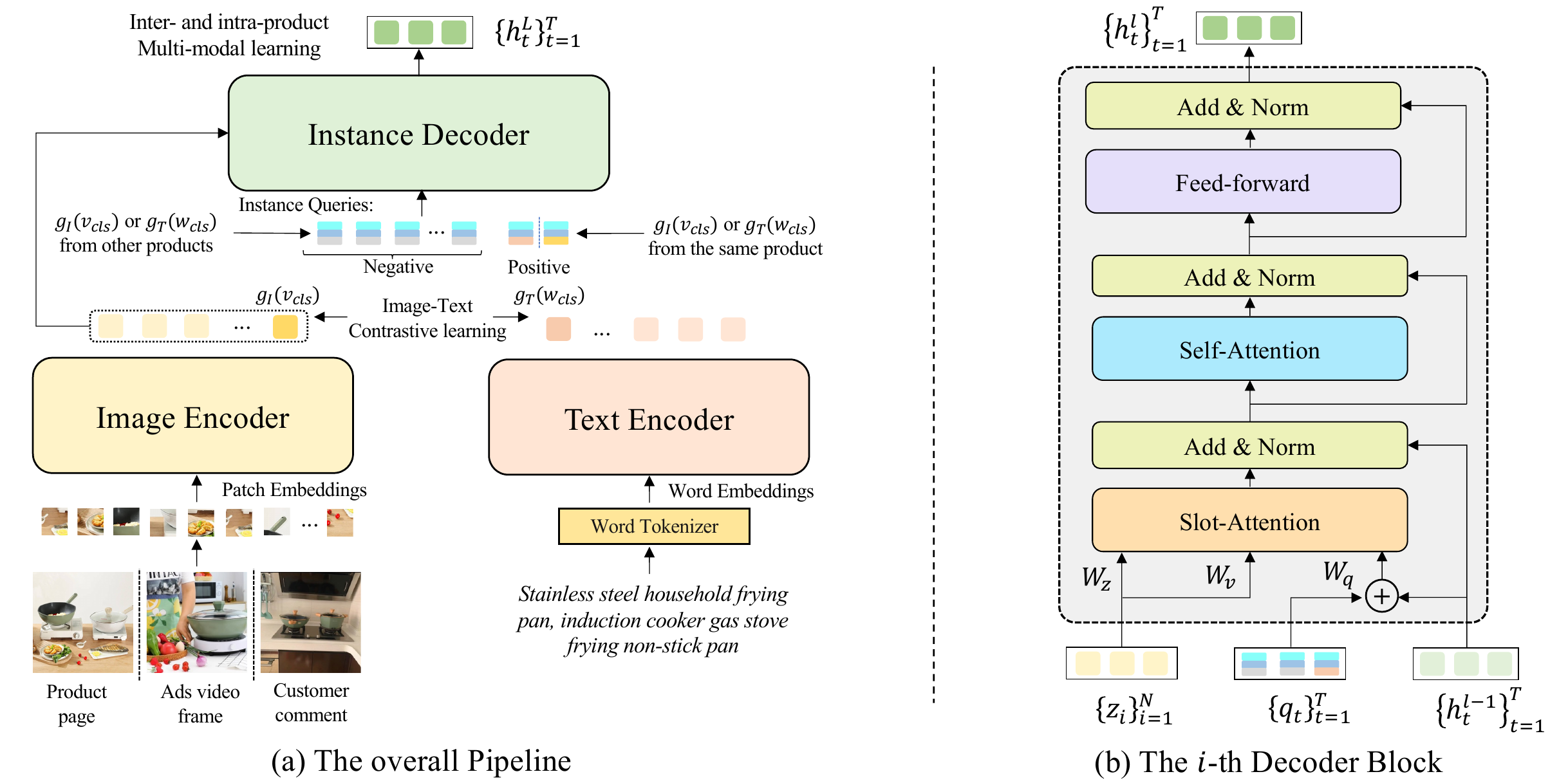}
  \vspace{-0.5em}
  \caption{(a) The architecture of the proposed instance-level representation learning paradigm (ECLIP), which is composed of an image encoder, a text encoder, and an instance decoder. The whole framework is optimized by three pretext tasks when pretraining on a large-scale E-commerce dataset. (b) The core decoder architecture that aims at aggregating the instance representation of the desired product. }
  \label{fig:fig2}
  \vspace{-1.6em}
\end{figure*}

\section{Approach}
\label{sec:method}

In this section, we begin with the overview of our proposed ECLIP in Section~\ref{sec:overview}. Then the core decoder architecture that aims at aggregating instance-level representation of the desired product is introduced in Section~\ref{sec:decoder}. To optimize the entire framework, we carefully designed several pretraining objectives in Section~\ref{sec:pretrain}. Finally, we delineate how to transfer the resulting foundation model to various downstream tasks in Section~\ref{sec:transfer}.

\subsection{Model Overview}
\label{sec:overview}
As illustrated in Figure~\ref{fig:fig2}, ECLIP is composed of an image encoder, a text encoder, and an instance decoder. Given an input sample $x=(x^I, x^T)$, where $x^I$ and $x^T$ are the image and text describing the corresponding product information, respectively. These two encoders first encode the image-text pair as a sequence of feature embeddings. Then, a modality-dependent projection layer is employed to map them linearly into a joint multi-modal feature space. These projected embeddings are further decoded to produce an instance-centric representation. Details of the two unimodal encoders are elaborated as follows.

\vspace{0.03in}

\noindent \textbf{Image Encoder}. 
Following the vision transformers~\cite{dosovitskiy2020image}, the product image $x^I \in \mathcal{R}^{H\times W \times C}$ is partitioned into $N$ non-overlapping patches. These patches are flattened to 1D input tokens and then projected linearly, with position embeddings added. Through hierarchical feature encoding, we can obtain a sequence of visual embeddings $\{ v_{cls}, v_1, ..., v_N \}$, where $v_{cls}$ indicates the special token $[\text{CLS}]$ for encoding the entire image information. 

\vspace{0.03in}

\noindent \textbf{Text Encoder}. 
This encoder adopts analogous transformer-style architecture. For the input product description $x_T$, it tokenizes the text to $M$ subwords as in BERT~\cite{devlin2018bert}. Similar to the image encoder, a special $[\text{CLS}]$ token is appended to the beginning of the textual input to summarize the text semantics. After encoding, the resulting linguistic embedding sequence is denoted as $\{ w_{cls}, w_1, ..., w_M \}$.

\subsection{Extract Instance-Centric Representation}
\label{sec:decoder}
After obtaining the contextualized embeddings, existing VLP approaches leverage $g_I(v_{cls}) \in \mathcal{R}^{D}$ and $g_T(w_{cls}) \in \mathcal{R}^{D}$ to align positive image-text pairs via contrastive learning. Here $g_I(\cdot)$ and $g_T(\cdot)$ are the aforementioned projections. While effective in the general domain, this design only considers the alignment between the global image-text semantics. However, in the E-commerce image, only several regions containing the desired product instance are informative foregrounds corresponding to the text description. Modeling such image-level alignment will fail to learn strong and robust product semantics. Hence, we are committed to learning instance-centric representation. 

\vspace{0.05in}

\noindent \textbf{Instance Query}. 
Motivated by~\cite{locatello2020object,carion2020end}, a set of learnable tokens called \textit{instance query} are introduced to ground the potential instance in the product image. As Figure~\ref{fig:fig2} shows, each query is correlated to a specific text or image that we refer to as \textit{multi-modal prompt}. The insight behind this design is that we wish the instance that a query should probe to be specified by the prompt content. Formally, the proposed instance query is denoted as $\bm{Q}=\{ q_t \in \mathcal{R}^{D}\}_{t=1}^{T}$, which can be obtained by:
\begin{equation}
    q_t = q_t^{\text{prompt}} + q_t^{\text{pos}} + q_t^{\text{type}}.
\end{equation}
Here, $q_t^{\text{prompt}}$ denotes $g_I(v_{cls})$ or $g_T(w_{cls})$, $q_t^{\text{pos}}$ and $q_t^{\text{type}}$ are learnable position and type embedding, indicating the probing area of a query and the modality type of the binding prompt. These queries are responsible for aggregating the instance-centric representations $\bm{H}=\{h_t\}_{t=1}^{T}$ from the encoded visual features via a decoder architecture. During pretraining, there is only one positive prompt (w.r.t the same product) for a given sample, and the rest $T-1$ are negative ones sampled from other products.

\vspace{0.05in}

\noindent \textbf{Instance Decoder}. 
We first project all the encoded $\{v_i\}_{i=1}^{N}$ into the same feature space as the prompt, yielding an embedding sequence $\bm{Z} = \{z_i \in \mathcal{R}^{D} \}_{i=1}^{N}$. Moreover, the instance representations $\bm{H}^0$ are zero-initialized, and then before feeding to the decoder. The proposed decoder then reads all the above-described embeddings: $\bm{Z}$, $\bm{Q}$ and $\bm{H}^0$ as its input. It has $L$ stacked blocks, and each one consists of a slot-attention and a self-attention layer.

The goal of the slot-attention layer is to adaptively update query representations through the interaction with the encoded visual embeddings. In detail, for the $l$-th slot-attention layer, it first calculates a similarity matrix $M \in \mathcal{R}^{N \times T}$, which is implemented by the dot-product attention mechanism~\cite{vaswani2017attention}. Formally, it is formulated by: 
\begin{equation}
    \label{eq:sim}
    M = \frac{1}{\sqrt{D}} (Z W_z) \cdot ((Q + H^{l-1}) W_q)^{\top},
\end{equation}
where $W_z$ and $W_q$ are the learnable projections parameter matrices, $\bm{H}^{l-1}$ is the instance representation produced by the $(l-1)^{\text{th}}$ decoder block. The similarity matrix $M$ is further normalized by a softmax function over $T$ queries:
\begin{equation}
    M_{ij} = \frac{\exp (M_{ij})}{\sum_{t=1}^{T} \exp (M_{it}) }.
\end{equation}
The generated matrix $M$ actually performs soft assignment via computing semantic similarity between $N$ visual tokens and $T$ instance queries. In doing so, it is capable of distributing each visual token to a specific query according to their similarity score. To aggregate the information of the visual tokens into their assigned input query, we compute a weighted mean update based on $M$:
\begin{equation}
\label{eq:update}
    \Delta h_t^{l-1} = \frac{1}{\sum_{i=1}^{N} M_{it}} \sum_{i=1}^{N} M_{it} (W_{v} z_i).
\end{equation}
Finally, the instance representation $\bm{H}^{l}$ at the $l$-th layer can be updated by a residual connection:
\begin{equation}
    h_t^{l} = h_t^{l-1} + W_o \Delta h_t^{l-1}.
\end{equation}
where $W_{v}$ and $W_{o}$ are linear transformation parameters. 

On the top of slot-attention layer, there is a self-attention module that performs information propagation between each query. In detail, given the previously updated $\bm{H}^{l}$, it employs standard multi-head self-attention (MSA) followed by a fully connected feed-forward network as in~\cite{vaswani2017attention}. After $L$ successive decoder blocks, we can obtain the final instance representation $\bm{H}^{L}$. It is noteworthy that, since the multi-modal prompts just participate in the similarity calculation in Eq.~\ref{eq:sim}, the resulting $\bm{H}^{L}$ thus contains only visual-modality information. 

\vspace{0.05in}

\noindent \textbf{Discussion}: The proposed decoder works like conducting a clustering on the image tokens, where each instance query serves as the centroid of a cluster. At each decoder block, it determines where each token belongs by measuring its distance from the centroids in the semantic space. The cluster centroid is then updated via a soft manner (Eq.\ref{eq:update}) based on the calculated distance. By stacking multiple decoder blocks, it can implicitly force each query to attend to a specific region and aggregate instance-level representations. 

\begin{figure}[t]
  \centering
  \includegraphics[width=0.85\linewidth]{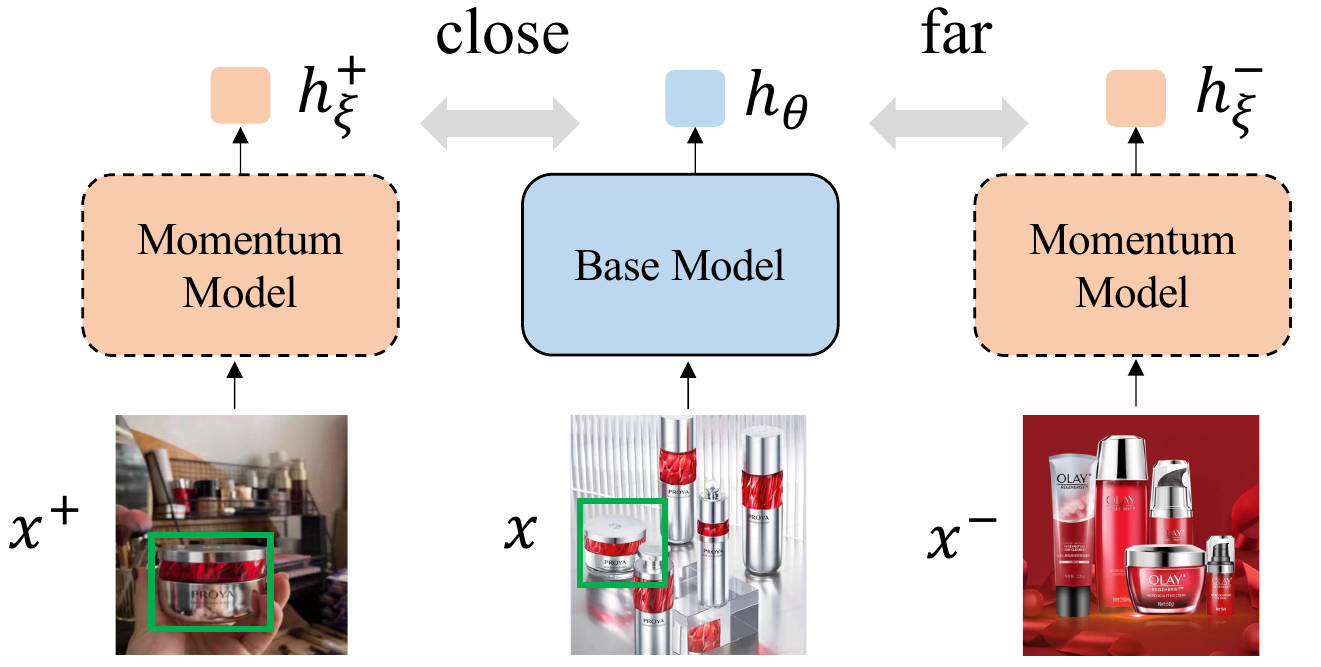}
  \caption{The illustration of inter-product contrastive learning. It aims to bring the positive samples from the same product ($x$ and $x^+$) closer than the negative ones ($x$ and $x^-$), which contributes to grounding the desired product (outlined with the green box).}
  \label{fig:fig3}
  \vspace{-1.5em}
\end{figure}

\subsection{Multi-Modal Pretraining Objectives}
\label{sec:pretrain}
Our ECLIP is optimized on large-scale uncurated product data with several pretraining proxy tasks. In the following, we describe each task in detail.

\vspace{0.05in}

\noindent \textbf{Image-Text Contrastive Learning}. 
As in~\cite{radford2021learning,jia2021scaling,li2021align}, this task contributes to learning better unimodal representations. Given a batch of product samples $\{(x_i^I, x_i^T)\}_{i=1}^B$, the similarity between image $x^I$ and text $x^T$ is estimated as:
\begin{equation*}
    s(x^I, x^T) = g_I(v_{cls})^\top g_T(w_{cls}).
\end{equation*}
This pretraining objective brings the image-text pairs of the same product in the embedding space closer than the unmatched ones, which consists of image-to-text term $\mathcal{L}_{i2t}$:
\begin{equation}
    \mathcal{L}_{i2t} = - \sum_{i=1}^{B}  \log \frac{\exp (s(x_i^I, x_i^T) / \tau)}{\sum_{j=1}^B \exp (s(x_i^I, x_j^T) / \tau)} ,
\end{equation}
and a text-to-image term $\mathcal{L}_{t2i}$:
\begin{equation}
    \mathcal{L}_{t2i} = - \sum_{i=1}^{B}  \log \frac{\exp (s(x_i^T, x_i^I) / \tau)}{\sum_{j=1}^B \exp (s(x_i^T, x_j^I) / \tau)},
\end{equation}
where $\tau$ is a learnable temperature parameter. The whole objective is then defined as $\mathcal{L}_{itc} = \frac{1}{2}(\mathcal{L}_{i2t} + \mathcal{L}_{t2i})$.

\vspace{0.05in}

\noindent \textbf{Inter-Product Multi-modal Learning}. 
As shown in Figure~\ref{fig:fig3}, we maintain a momentum model during pretraining that is an exponential-moving-average of the origin model like~\cite{he2020momentum}. For a product sample $x_i$, we denote the representation of the positive prompt produced by the base and momentum model as $h_{\theta}^i$ and $h_{\xi}^i$. An inter-product contrastive loss $\mathcal{L}_{inter}$ is computed by:
\begin{small}
\begin{equation}
\mathcal{L}_{inter} = - \sum_{i=1}^{B}  \log \frac{\exp ({h_{\theta}^i}^\top h_{\xi}^j / \tau)}{\exp ( {h_{\theta}^i}^\top h_{\xi}^j / \tau ) + \sum_{k \in \mathcal{N}^{-}} \exp ({h_{\theta}^i}^\top h_{\xi}^k / \tau)},
\end{equation}
\end{small}
where sample $i$ and $j$ are a positive pair, $\mathcal{N}^{-}$ is a negative sample set that belongs to other products. This objective maximizes the similarity between different samples of the identical product, while minimizing those of the unmatched ones. Since the images of a product are collected from different sources, their background appearances are usually diverse. Hence, $\mathcal{L}_{inter}$ will encourage the yielded representation to be highly correlated with the desired product and thus contributes to aligning the positive prompt with the corresponding image tokens in a fine-grained manner.

This pretext task also incorporates an additional instance-text matching loss that predicts whether an instance and a text description are matched. Formally, given an instance-text pair, we obtain their match logit defined by: $f(h_\theta^i \odot g_T(w_{\text{cls}}^i))$, where $\odot$ is Hadamard product and $f(\cdot)$ is a mapping layer: $\mathcal{R}^D \rightarrow \mathcal{R}^2$. This match logit is optimized by a typical binary cross entropy loss $\mathcal{L}_{itm}$. 

\vspace{0.05in}

\noindent \textbf{Intra-Product Multi-modal Learning}. 
For a product sample, there is only one positive prompt describing the presented product during pretraining, and the rest $T-1$ prompts are sampled from other products. The motivation behind this pretext task is to ensure that only positive queries can probe the foreground instance, rather than negative ones. To this end, we apply an intra-product contrastive loss using text supervision. Suppose index $r$ indicates the positive query, then $\mathcal{L}_{intra}$ can be formulated as:
\begin{equation}
    \mathcal{L}_{intra} = - \sum_{i=1}^{B}  \log \frac{\exp ({h_r^i}^\top g_T(w_\text{cls}^i) / \tau)}{\sum_{t=1}^T \exp ({h_t^i}^\top g_T(w_\text{cls}^i) / \tau)},
\end{equation}
which serves to bring the positive query and the product description closer than all $T-1$ negative ones. Moreover, we also introduce an entropy regularization term for $M$:
\begin{equation}
\begin{aligned}
    \mathcal{L}_{\mathcal{R}} = \sum_{i=1}^{N} & M_{i,r} \log ( \frac{1}{M_{i,r}} ) + \\
    & \sum_{j=1, j \neq r}^{T} \left (\log N - \sum_{i=1}^{N} M_{i,j} \log ( \frac{1}{M_{i,j}} ) \right ).
\end{aligned}
\end{equation}
This regularization term encourages the positive query to focus on several tokens that may contain product instances. While for $T-1$ negative ones, it prevents too sharp similarity distributions over $N$ image tokens. Finally, the overall pretraining objective of ECLIP is the sum of all aforementioned loss terms.

\subsection{Transfer to Downstream Tasks}
\label{sec:transfer}
Once pretrained, the resulting foundation model can be leveraged to extract the product instance representation with minimal surgery. Specifically, given a product sample $(x_i^I, x_i^T)$, we first encode the image-text pair into an embedding sequence via the unimodal encoders. Then, the global representation of text description $g_T(w_\text{cls})$ is treated as the positive query and fed into the decoder concatenated with $T-1$ negative ones. Here, the negative queries $\{q_t\}_{t=2}^{T}$ are sampled from a standard Gaussian distribution for convenience. We also explore different negative query setting manners in Section~\ref{sec:abla}. The yielded representation $h_0^{L}$ belonging to the positive query is then applied to a wide range of E-commerce downstream tasks.

\section{Experiments}
\label{sec:exp}
\subsection{Pretraining Details}
\noindent \textbf{Pretraining Dataset}
We collected a large-scale pretraining dataset from a popular E-commerce website. It consists of 15M different products and over 100M various images, covering about 9K diverse categories such as clothes, daily necessities, instruments, and so on. For each product item, it has a corresponding textual description and several images from the product details pages, customer comments, and attached advertisement videos. During pretraining, the positive data pairs are constructed by sampling images belonging to the same product from different sources.

\noindent \textbf{Implementation Details}. 
The image encoder adopts the same network configuration as the standard ViT~\cite{dosovitskiy2020image} and is initialized from the weight pre-trained on ImageNet. Our text encoder is implemented with the same architecture as $\text{BERT}_{\text{base}}$~\cite{devlin2018bert}. The decoder has 6 identical blocks and 20 instance queries. We here explore two variants of ViTs: ViT-B/16 and ViT-L/16. There is a total of 220 / 450 million parameters for the base and large version. During pretraining, the input images are resized to 224 $\times$ 224 with random crop and horizontal flip augmentation, and the texts are tokenized by WordPiece with a maximum length of 55. We pretrain for 15 epochs with a batch size of 6400 (ViT-B) / 4096 (ViT-L) on 32 NVIDIA A100 GPUs. The whole framework is learned using AdamW~\cite{loshchilov2017decoupled} optimizer and the learning rate is warmed up to 1e-4 and then decayed linearly. More details are elaborated in the supplementary.

\noindent \textbf{Compared Baselines}. 
We mainly compare the ECLIP with several state-of-the-art VLP methods: CLIP~\cite{radford2021learning}, FILIP~\cite{yao2021filip}, DeCLIP~\cite{li2021supervision}, ALBEF~\cite{li2021align} and BLIP~\cite{li2022blip}. For a fair comparison, we also leverage ViT-B/16 as the image encoder, $\text{BERT}_{\text{base}}$ as the text encoder, and pretrain these baselines on the same 100M E-commerce data using their official public implementations. 

\subsection{Evaluation on Downstream Tasks}
Next, we delineate the evaluation performances for five specific E-commerce downstream tasks in turn.

\vspace{-0.2in}

\subsubsection{Zero-Shot Product Classification}
We first transfer ECLIP to product item classification. It is a recognition task that aims to map a product sample to a specific category. We evaluate the performance on a large-scale publicly available E-commerce dataset called M5Product~\cite{dong2022m5product}, which covers 1.1M images and 5,679 various product categories. Here, we consider the multimodal setting, which uses both the product image and related textual description for classification. To demonstrate the strong zero-shot ability of ECLIP, we apply it directly to the classification evaluation without further finetune. It is achieved by measuring the similarity between the category text like CLIP~\cite{radford2021learning}. The left part of Table~\ref{tab:classify_retrieval} summarizes the Top-$1$ classification accuracy of all the compared methods. As presented, our ECLIP greatly exceeds all the existing baselines by a large margin (e.g., +6.6\% v.s. CLIP), demonstrating the superiority of instance-level representation. 

\vspace{-0.1in}

\subsubsection{Zero-Shot Image-Text Retrieval}
ECLIP is also transferred to test zero-shot performance for image-to-text and text-to-image retrieval. To this end, we collect a large dataset that contains 205K image-text pairs of E-commerce products. Since only unimodal information is available in this task, we simply use our image and text encoders to embed image-text pairs and complete the retrieval based on their pairwise similarity. We utilize the widely-used Recall$@K$ metric for evaluation. Detailed comparison results are shown in the right part of Table~\ref{tab:classify_retrieval}. We can see that, despite training on the same dataset, our method achieves superior performance owing to fine-grained alignment modeling between text and product instance.

\begin{table}[h]
    \centering
    \resizebox{\linewidth}{!}{
    \begin{tabular}{lccccc}
    \toprule
    \multirow{2}{*}{Method} & Classification  & \multicolumn{2}{c}{Image-to-Text} & \multicolumn{2}{c}{Text-to-Image} \\
    \cmidrule(lr){2-2} \cmidrule(lr){3-4}  \cmidrule(lr){5-6} &  Acc@1  & R@1 & R@5 & R@1 & R@5 \\
    \midrule
    CLIP~\cite{radford2021learning} & 37.2   & 52.6  & 74.1 & 58.7 & 84.0 \\
    FILIP~\cite{yao2021filip} &  37.1 &  52.3    & 73.8  & 58.0 & 83.5\\
    DeCLIP~\cite{li2021supervision}  & 37.8   &  53.1  & 75.8 & 58.8 & 83.9 \\
    ALBEF~\cite{li2021align} &   38.5    &  52.9  & 74.4 & 58.2 & 83.3 \\
    BLIP~\cite{li2022blip} &   39.3     &  53.3 & 75.6  & 59.1 & 84.4 \\
    $\text{Ours}_{\text{ViT-B/16}}$  &  43.8  & 53.8  & 76.0 & 59.9 & 84.6  \\
    $\text{Ours}_{\text{ViT-L/16}}$  & \textbf{44.8}   & \textbf{58.2}  & \textbf{79.6} & \textbf{63.8} & \textbf{87.4} \\
    \bottomrule
    \end{tabular}
    }
    \caption{Performance comparisons of zero-shot product classification and zero-shot image-text retrieval on M5Product dataset~\cite{dong2022m5product}.}
    \label{tab:classify_retrieval}
    \vspace{-1em}
\end{table}

\vspace{-1em}

\begin{table*}
    \centering
    \resizebox{\linewidth}{!}{
    \begin{tabular}{lcccccccccc}
    \toprule
    \multirow{2}{*}{Method} & \multirow{2}{*}{Pretraining Dataset} & \multicolumn{3}{c}{Coarse Product Retrieval} & \multicolumn{6}{c}{Fine-grained Product Retrieval} \\
    \cmidrule(lr){3-5} \cmidrule(lr){6-11} & & mAP@1 & mAP@5 & mAP@10 & R@1 & R@5 & R@10 & mAP@1 & mAP@5 & mAP@10 \\
    \midrule
    ViLBERT~\cite{lu2019vilbert} & \multirow{3}{*}{M5Product} & 58.6  &  61.7 &  60.1 &  - & -  & - & -  & -  & - \\
    UNITER~\cite{chen2020uniter} & & 58.9  &  62.8 &  60.9 &  - &  - & -  & -  &  - &  - \\
    SCALE~\cite{dong2022m5product} & & 59.8  & 64.1 & 62.2  & -  & -  &  - & -  & -  & -   \\
    \midrule
    CLIP~\cite{radford2021learning} & \multirow{7}{*}{ECLIP 100M} & 68.2 & 73.2  & 70.7  & 34.8 & 54.2 & 62.9 & 34.8 & 40.2 & 39.9 \\
    FILIP~\cite{yao2021filip} & &  67.8   & 73.0 & 70.3  & 34.6  & 53.9  & 62.2  & 34.6 &  40.1 & 39.7 \\
    DeCLIP~\cite{li2021supervision} & & 68.5  & 73.4 & 70.8 & 35.3 & 56.4 & 65.5 & 35.3 & 41.2 & 40.8 \\
    ALBEF~\cite{li2021align} & &   68.7 & 73.6 & 71.2 & 35.1 & 56.1 & 65.2 & 35.1 & 40.7 & 40.4  \\
    BLIP~\cite{li2022blip} & &  69.1 & 74.1  & 71.6 & 35.6 & 56.8 & 66.0 & 35.6 & 41.6 & 41.3 \\
    $\text{Ours}_{\text{ViT-B/16}}$ & & 69.6 & 74.9 & 72.5 &  44.3  & 63.4 &  71.1 & 43.8 & 48.6 & 48.2  \\
    $\text{Ours}_{\text{ViT-L/16}}$ &  & \textbf{70.2}  & \textbf{75.3} &  \textbf{72.9}  & \textbf{45.0} & \textbf{64.2} & \textbf{72.1} & \textbf{45.0} & \textbf{50.0} & \textbf{49.5} \\
    \bottomrule
    \end{tabular}
    }
    \caption{Performance comparisons of zero-shot coarse level and fine-grained level product retrieval task.}
    \label{tab:product_retrieval}
    \vspace{-1em}
\end{table*}

\subsubsection{Zero-Shot Product Retrieval}
This task aims to find the most relevant target product given a query (image-text pair of a product). It has a wide range of applications in real e-commerce scenarios such as recommending relevant products for users. We first evaluate the coarse-level retrieval. Following~\cite{dong2022m5product}, a product pair is considered a match if both belong to the same category during evaluation. The results on M5Product benchmark are reported on the left of Table~\ref{tab:product_retrieval}. It can be seen that exploiting instance-centric representation significantly boosts performance. To further demonstrate the effectiveness of instance-level representation, we then conduct a more complicated fine-grained level product retrieval task, where a pair is considered a match if and only if they are the same product. This task requires more adequate fine-grained understanding ability since it focuses on the specific product instance. Detailed comparison results are shown on the right part of Table~\ref{tab:product_retrieval}. One can find that our ECLIP achieves a substantial improvement in retrieval performance (e.g., 44.3\% v.s. 35.6\% (BLIP) on R@1). 

We also consider another setting introduced in~\cite{zhan2021product1m}, called instance-level retrieval, where the query image encompasses multiple different kinds of product instances. And the model needs to find all the related products from a large gallery. As shown in Table~\ref{tab:instance_retrieval}, ECLIP still achieves superior performance than all the previous approaches. Although the CAPTURE leverages a specially-trained object detector to extract instances, ECLIP still surpasses it by a clear margin with no box annotation.

\vspace{-0.1in}

\subsubsection{Zero-Shot Visual Grounding}
To demonstrate whether our model possesses the capability of localizing the desired product instance after pretraining, we further evaluate ECLIP on the zero-shot product grounding, which requires localizing the product instance in an image according to a textual description. Specifically, the input image-text pair is first fed to our ECLIP to obtain a score map $\mathcal{S} \in \mathcal{R}^{H \times W}$ that measures the similarity between the text and each image location. Then, we use $\mathcal{S}$ to rank the candidate regions produced by an off-the-shelf region proposal network. The performance is evaluated by the top-1 accuracy at IoU thresholds $\{0.5, 0.7\}$ on an annotated grounding dataset consisting of 450K product images. Detailed comparison results are listed in Table~\ref{tab:grounding}. As we can see, compared to methods aimed at global representation, our model has learned fine-grained cross-modal understanding ability and thus obtains substantial performance gain (e.g., +14.5\% v.s. BLIP on Acc@0.7). Since ECLIP supports image prompts during pretraining, we also conduct zero-shot image-conditioned grounding. The results and analysis can be found in the supplementary.

\vspace{-0.1in}
\subsubsection{Transfer to Object Detection}
\begin{figure*}
  \centering
  \begin{subfigure}{0.50\linewidth}
    \includegraphics[width=0.95\linewidth]{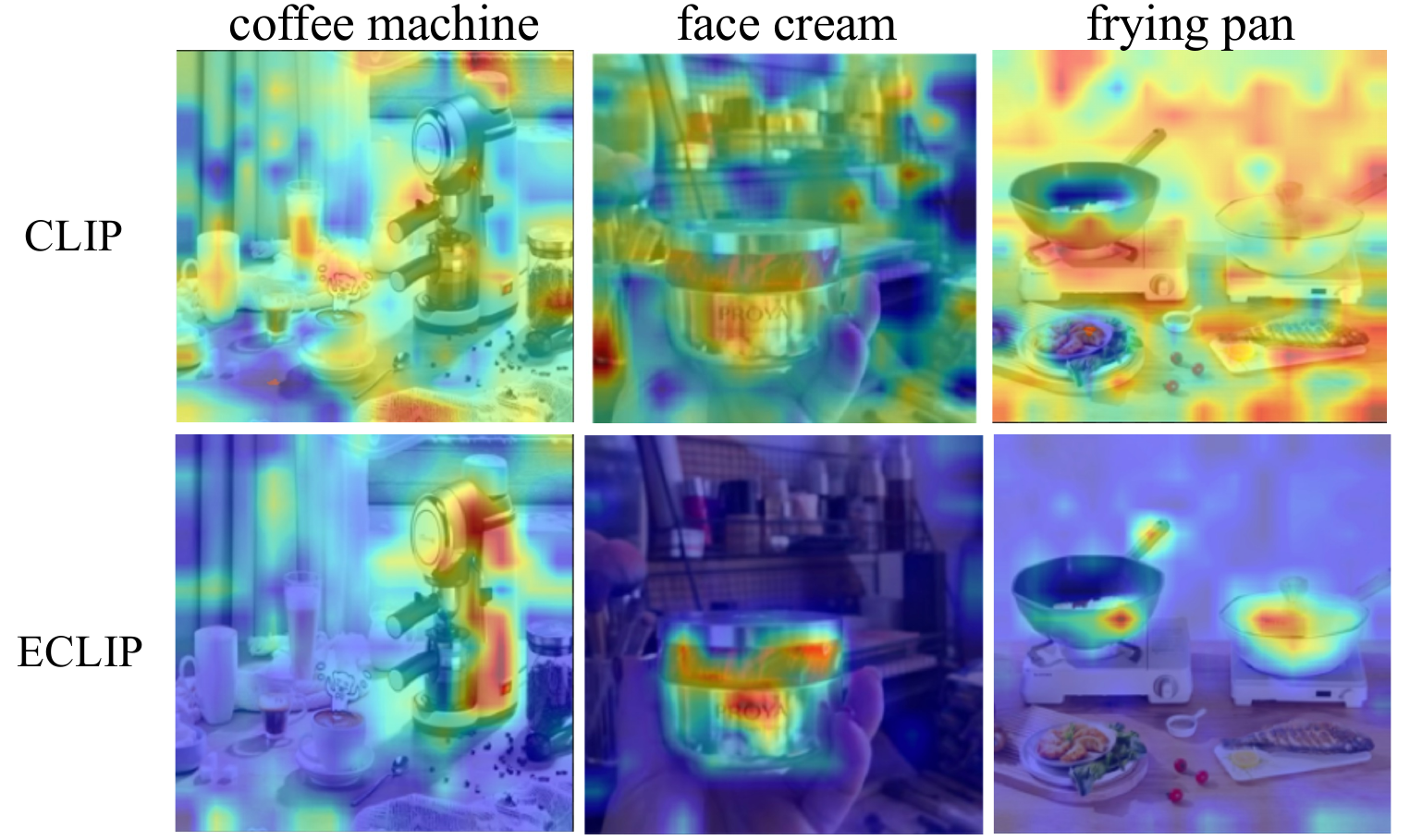}
    \caption{Visualization of zero-shot visual grounding.}
    \label{fig:attn}
  \end{subfigure}
  \hfill
  \begin{subfigure}{0.46\linewidth}
    \includegraphics[width=0.88\linewidth]{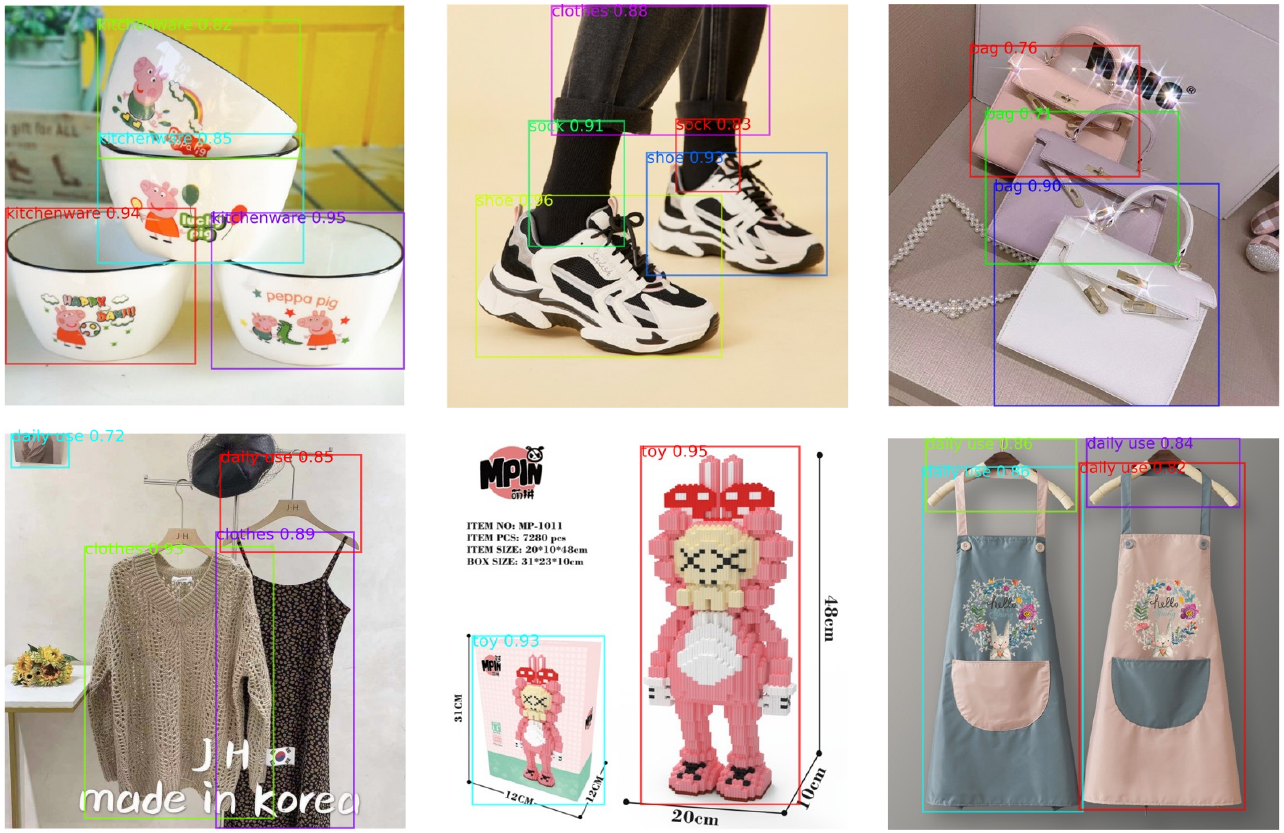}
    \caption{Object detection examples after finetune.}
    \label{fig:pred_example}
  \end{subfigure}
  \caption{Qualitative examples of visual grounding and object detection. See the main text for more explanation.}
  \label{fig:vis}
  \vspace{-1em}
\end{figure*}

We also transfer ECLIP to object detection to further validate its fine-grained understanding ability. Following DETR~\cite{carion2020end}, we utilize the image encoder to embed visual features, and utilize the decoder with a newly added prediction head to decode the potential objects. Moreover, we collect a manually annotated detection dataset covering 160K images. We split a 20K subset for evaluation and leave the rest for model finetuning. The supplementary provides experiment details of baselines and ECLIP. It can be observed from Table~\ref{tab:grounding} and Figure~\ref{fig:pred_example} that it outperforms the existing VLP methods, demonstrating the superiority of ECLIP to learning fine-grained semantics in E-commerce. 

\begin{table}[t]
    \centering
    \resizebox{\linewidth}{!}{
    \begin{tabular}{lcccc}
    \toprule
    \multirow{2}{*}{Method}  & \multicolumn{4}{c}{Instance-Level Product Retrieval}  \\
    \cmidrule(lr){2-5}  &  mAP@10  & mAP@50  & mAR@10  & mAR@50 \\
    \midrule
    CAPTURE~\cite{zhan2021product1m} & 40.4   &  36.8 &  17.2 & 15.9   \\
    CLIP~\cite{radford2021learning} &  86.6  & 82.8  & 54.4 &  59.5  \\
    FILIP~\cite{yao2021filip} &  86.9 & 83.0   &  54.6  & 59.8  \\
    DeCLIP~\cite{li2021supervision}  &  87.1  & 83.3   & 54.9 & 60.0   \\
    BLIP~\cite{li2022blip} & 87.5 &  83.5 &  55.1 & 60.4 \\
    $\text{Ours}_{\text{ViT-B/16}}$  &  \textbf{89.6}  & 84.6  & 55.9 &  61.2  \\
    $\text{Ours}_{\text{ViT-L/16}}$  & 89.5  &  \textbf{86.4} & \textbf{56.3} & \textbf{62.1} \\
    \bottomrule
    \end{tabular}
    }
    \vspace{-0.5em}
    \caption{Performance comparisons of zero-shot instance-level product retrieval task on Product1M dataset~\cite{zhan2021product1m}.}
    \label{tab:instance_retrieval}
    \vspace{-0.7em}
\end{table}

\begin{table}[t]
    \centering
    \resizebox{\linewidth}{!}{
    \begin{tabular}{lcccc}
    \toprule
    \multirow{2}{*}{Method} & \multicolumn{2}{c}{Visual Grounding} & \multicolumn{2}{c}{Object Detection} \\
    \cmidrule(lr){2-3} \cmidrule(lr){4-5} & Acc@0.5 & Acc@0.7   & mAP@0.3 & mAP@0.5  \\
    \midrule
    CLIP~\cite{radford2021learning}   &  80.9 & 75.2 & 17.2 & 14.3 \\
    FILIP~\cite{yao2021filip}  &    81.3  & 75.6  & 17.5 & 14.6 \\
    DeCLIP~\cite{li2021supervision}     &  81.0  & 75.3 & 17.0 & 14.2  \\
    ALBEF~\cite{li2021align}     &  80.9  & 74.7 & 17.1 & 13.9 \\
    BLIP~\cite{li2022blip}      &  81.1 & 75.1  & 17.3 & 14.1 \\
    $\text{Ours}_{\text{ViT-B/16}}$   & \textbf{91.2}  & \textbf{89.6} & \textbf{20.2} & \textbf{16.5}  \\
    \bottomrule
    \end{tabular}
    }
    \vspace{-0.5em}
    \caption{Performance comparisons of zero-shot visual grounding and finetuned object detection. }
    \label{tab:grounding}
    \vspace{-1.7em}
\end{table}

\subsection{Ablation Study}
\label{sec:abla}
\noindent \textbf{Effect of Pretext Task}. 
To demonstrate the effectiveness of the inter- and intra- product learning pretext task, we designed the experiments on different task combinations for the product retrieval task. All the ablations were conducted on a smaller pretraining dataset that includes only 5M images due to the costly training time. The complete results are listed in Table~\ref{tab:pretext}. One can observe that canceling either of these two tasks will result in worse performance. Notably, the inter-product task brings a more significant performance boost compared to the intra-product. We speculate that because the former contrasts with more negative samples from different product images.

\noindent \textbf{Effect of Negative Query}. 
Since the negative instance queries are used when transferring to classification and retrieval tasks, we also ran ablations on different ways of setting these negative ones. We tried the following cases: 1) Leverage the description text from other products and encode with the text encoder. 2) Randomly sample the negative queries from the standard Gaussian distribution. 3) Adopt the exponential moving average of queries on the whole dataset during pretraining. Table~\ref{tab:neg_query} summarizes the results of this ablation study on the product retrieval task. As observed, there is little to no difference between different manners of setting up negative queries. We thus adopt random sampling for implementation simplicity.

\begin{table}[h]
\centering
\resizebox{0.75\linewidth}{!}{
\begin{tabular}{cc|cc}
\toprule
$\mathcal{L}_{inter}$ &  $\mathcal{L}_{intra}$ & mAP@1 (\%) & mAP@5 (\%) \\
\midrule
&           &  61.6   &  65.2   \\
\checkmark &    &  64.4  &  68.7  \\
 & \checkmark   &  62.2   &  66.5 \\
\checkmark & \checkmark  & \textbf{65.2} &  \textbf{69.3} \\
\bottomrule
\end{tabular}}
\vspace{-0.5em}
\caption{Ablation of different pretraining pretext task combinations on coarse-level product retrieval task (ViT-B/16). }
\label{tab:pretext}
\vspace{-1em}
\end{table}

\vspace{-0.5em}

\begin{table}[h]
\centering
\resizebox{0.75\linewidth}{!}{
\begin{tabular}{c|cc}
\toprule
Metric  & mAP@10  &  mAR@10  \\
\midrule
Negative Text &   84.9   & 52.5 \\
EMA Update &  85.1  & 53.3 \\
Random Sampling & \textbf{86.8} & \textbf{54.1} \\

\bottomrule
\end{tabular}}
\vspace{-0.5em}
\caption{Ablation of different negative query setting manners on the instance-level product retrieval task (ViT-B/16).}
\label{tab:neg_query}
\vspace{-1em}
\end{table}

\subsection{Qualitative Analysis}
\begin{figure}[t]
  \centering
  \includegraphics[width=1\linewidth]{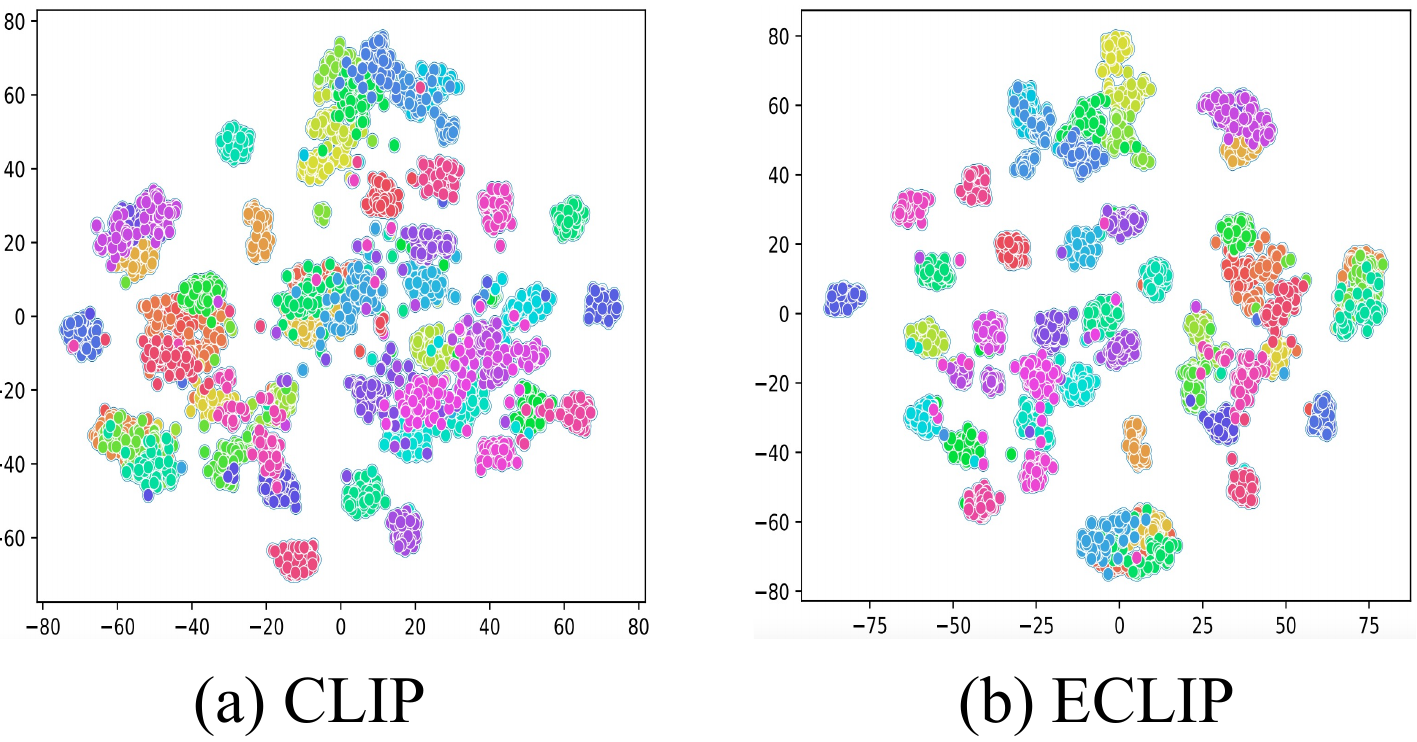}
  \vspace{-1.7em}
  \caption{The T-SNE visualization for the learned representations of CLIP and our ECLIP. Both of them are pretrained on the same dataset. We first randomly select 50 kinds of products and then draw 100 samples for each. Points belonging to the same category are of the same color. See the main text for more explanation.}
  \label{fig:tse}
  \vspace{-1.5em}
\end{figure}

In this section, we first qualitatively showcase that ECLIP can learn fine-grained cross-modal alignment to ground the desired product. Figure~\ref{fig:attn} presents the visualization of the similarity score map between a product image and its corresponding text description, where darker color indicates image locations with higher similarity to text. We can clearly observe that our model can rightly attend to the desired instance depicted by the text. Furthermore, T-SNE is used for visualizing the visual embedding of different kinds of product samples. As shown in Figure~\ref{fig:tse}, compared to CLIP, our ECLIP can extract semantic-rich yet compact representations that better distinguish products belonging to different categories. More visualization examples and analysis are provided in our supplementary.


\section{Conclusion}
In this paper, we develop an effective large-scale multi-modal pretraining paradigm called ECLIP in E-commerce. Beyond regular global representation, it instead aims to learn the instance-level representation via a novel decoder and the carefully-designed pretraining proxy tasks. Extensive experimental results further demonstrate the superior generalization capability of the proposed framework. 

{\noindent \small \textbf{Acknowledgement}: This work is supported by National Key R\&D Program of China (2022ZD0160305) and
Beijing Natural Science Foundation (Z190001).}

{\small
\bibliographystyle{ieee_fullname}
\bibliography{egbib}
}

\newpage

\appendix

\section*{Appendix}

In this supplementary material, we first present more implementation details about the pretraining dataset and model architecture in Section~\ref{sec:supp_imp}. Then, more experimental details and results analysis on the downstream tasks are given in Section~\ref{sec:supp_exp}. To better demonstrate the superior generalization and grounding capability of ECLIP, we illustrate more visualization examples in Section~\ref{sec:supp_vis}. Finally, additional analysis and discussion are provided in Section~\ref{sec:supp_board}. 

\section{More Pre-training Details}
\label{sec:supp_imp}
\subsection{Pre-training Dataset Details}
A large-scale dataset is indispensable for training a powerful foundation model. To this end, we construct a massive E-commerce pretraining dataset that consists of 100M image-text pairs and includes various product categories. All data samples are collected directly from a popular E-commerce website without further manual annotation. The details of the dataset collection are elaborated as follows:

\vspace{0.02in}

\noindent \textbf{Sample various products}. We first collect a large number of products covering various categories: clothes, daily use, cosmetics, etc. To avoid long-tail distribution, these products are further uniformly sampled according to their categories. After processing, we harvest around 12M product items covering a total of 9K categories.

\vspace{0.02in}

\begin{figure}[t]
  \centering
  \includegraphics[width=0.95\linewidth]{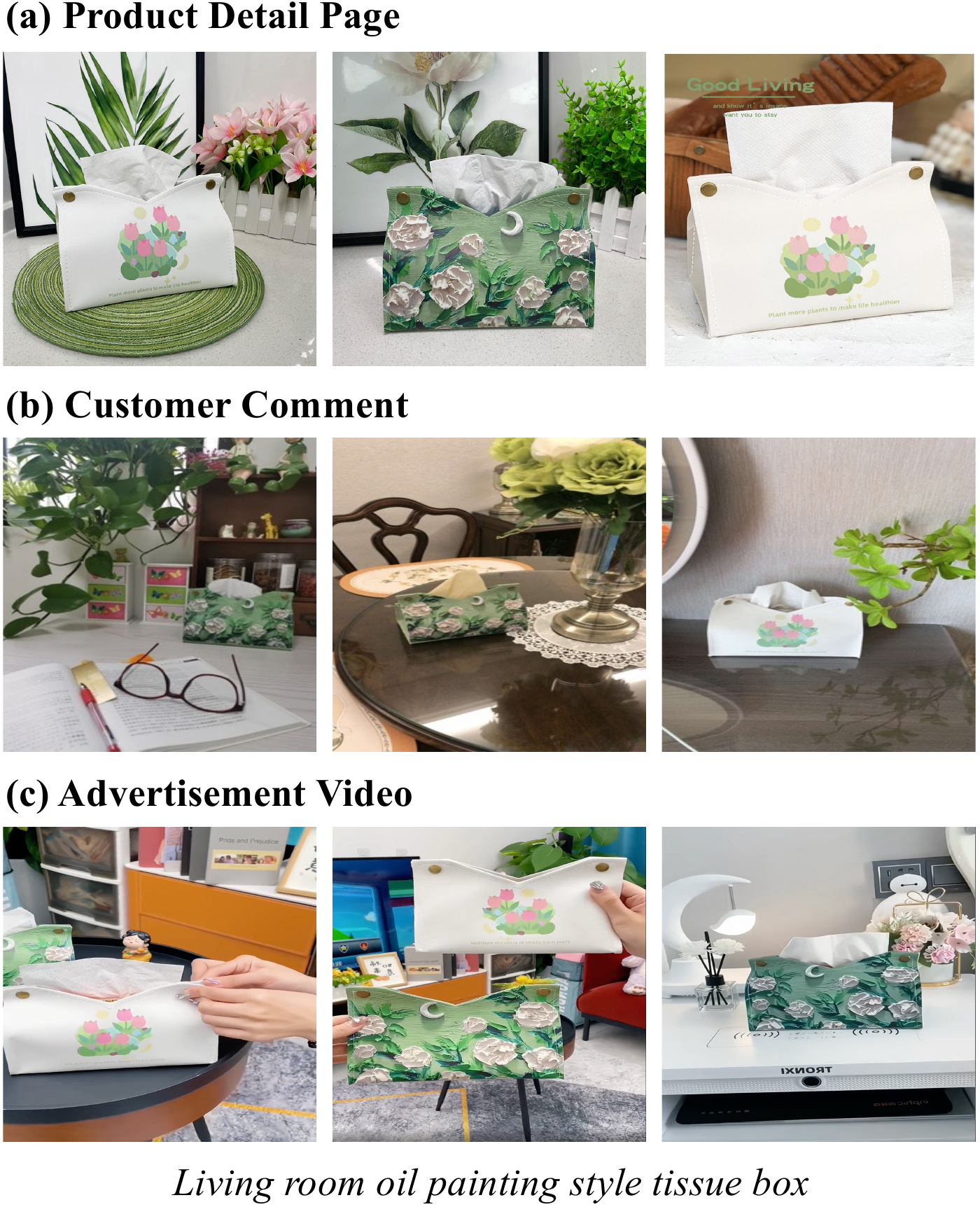}
  \vspace{-0.5em}
  \caption{Images of one product from different sources.}
  \label{fig:supp_fig1}
  \vspace{-1.7em}
\end{figure}

\noindent \textbf{Collect images from different sources}. For each collected product item, we sample several image samples from different sources: product details pages, customer comments, and attached advertisement videos. As shown in Figure~\ref{fig:supp_fig1}, the detail pages contain 3-4 images provided by merchants to display the product being sold from multiple views. We also select the 3 images taken by customers from the hottest comments related to a product item. In addition, the attached advertisement videos showcase the product’s appearance and usage to customers, and we randomly sample 5-6 video frames as image samples. For product items without comments and advertisement videos, we only collect images from product details pages. In summary, there are about 100M diverse E-commerce images.

\vspace{0.02in}

\noindent \textbf{Make positive training pairs}. During pretraining, two image samples from different sources but belonging to the identical product are treated as positive pairs. Both positive image samples have the same text description. For the few items with less than two images, we treat their randomly augmented image as the positive pair. In each training batch, we ensure that only two images are from the identical product, and the others are from different ones.

\subsection{Implementation Details}
In our instance decoder, the head number of the self-attention module is 8 and the hidden dimension of the feed-forward layers is 2048. The common embedding dim $D$ is 512 for all variants. An AdamW optimizer~\cite{loshchilov2017decoupled} is used for training with weight decay $0.01$. The learning rate for image and text encoders is set to $5e^{-5}$, and $1e^{-4}$ for the rest modules, which is first linearly warmed up and then exponentially decayed with a factor of $0.85$. Besides, we set the exponential-moving-average update parameter to $0.998$. Inspired by~\cite{li2021align}, we sample hard negative prompts based on global contrastive image-to-text and text-to-image similarities. This strategy contributes to learning a stronger and more robust representation. Following~\cite{he2020momentum}, the queue size for inter-product contrastive learning is set to 65,536. The overall pretraining procedures of ECLIP are two-stage: first freeze the weight of the instance decoder and only train two encoders with $\mathcal{L}_{itc}$ for 10 epochs. Then, train the whole components with all loss terms for another 5 epochs. In the second stage, the gradients of decoders are not propagated to two encoders. 

\section{More Experimental Details and Results}
\label{sec:supp_exp}
Additional details for transferring ECLIP to various downstream E-commerce tasks are described below.

\subsection{Downstream Task Details}

\subsubsection{Zero-Shot Product Retrieval}
For coarse-level product retrieval, the query and gallery set contains 24,410 and 1,197,905 product samples following~\cite{dong2022m5product}. During the evaluation, a product pair is considered a match if both belong to the same category. For instance-level retrieval, we follow the setting introduced in~\cite{zhan2021product1m}: each query image encompasses multiple different kinds of product instances and the model needs to retrieve all the related products. There are 9,220 query samples and 40,033 gallery samples in the instance-level retrieval. Similar to the classification task, we conduct retrieval using the image-text pair of a product. It is worth noting that the text descriptions between the matched query and gallery samples in M5Product and Product1M benchmarks are very similar. In this case, text modality will dominate the retrieval performance, which is harmful to reflecting the role of visual features. Therefore, for the challenging fine-grained product retrieval, we build a dataset that contains 26,000 products as the query set and 130,000 products as the gallery set, where the text descriptions of matched pairs are adequately different. Besides, a product pair is considered a match if and only if they are the same products. The significance of instance features can be fully verified in this task.

\subsubsection{Zero-Shot Visual Grounding}
\label{sec:ground}
Our ECLIP learns the fine-grained cross-modal alignment capability to localize the desired instance indicated by the text. To validate this, we transfer ECLIP to zero-shot visual grounding task. The collected grounding dataset has 484,385 image-text pairs, where each image contains only one ground truth box corresponding to the textual description. To evaluate the grounding performance, following~\cite{yu2018mattnet}, we first leverage an off-the-shelf region proposal network to extract a set of bounding-box proposals for each image. This network is pretrained on a human-annotated object detection dataset. Then, we estimate the similarity score between the text and each $16 \times 16$ image patch. The obtained 2D score map is further interpolated to the origin input resolution: $\mathcal{S} \in \mathcal{R}^{H \times W}$. Each box proposal $b=(x1,y1,x2,y2)$ is ranked based on $r(\cdot)$:
\begin{equation}
   r(b) = \frac{1}{\sqrt{\text{area}(b)}} \sum_{x=x_1}^{x2} \sum_{y=y_1}^{y2} S_{x,y}
\end{equation}
We select box with the maximum $r(\cdot)$ as the grounding result. The performance is finally evaluated by accuracy at IoU thresholds $\{0.5, 0.7\}$ with the ground truth box.

\begin{table}[t]
    \centering
    \resizebox{0.8\linewidth}{!}{
    \begin{tabular}{llcc}
    \toprule
    \multirow{2}{*}{Setting} & \multirow{2}{*}{Methods} & \multicolumn{2}{c}{IoU Thresh} \\
    \cmidrule{3-4} &   &   Acc@0.5  & Acc@0.7 \\
    \midrule
    \multirow{3}{*}{Zero-Shot} & CLIP~\cite{radford2021learning} &  79.8  &  72.1  \\
    & ALBEF~\cite{li2021align} & 80.2  &   74.8  \\
    & $\text{Ours}_{\text{ViT-B/16}}$  &  \textbf{88.7} &  \textbf{85.6}   \\
    \bottomrule
    \end{tabular}}
    \caption{Performance comparisons of different approaches to zero-shot image-conditioned visual grounding.}
    \label{tab:supp_grounding}
    \vspace{-1em}
\end{table}

\begin{figure*}[t]
  \centering
  \includegraphics[width=0.92\linewidth]{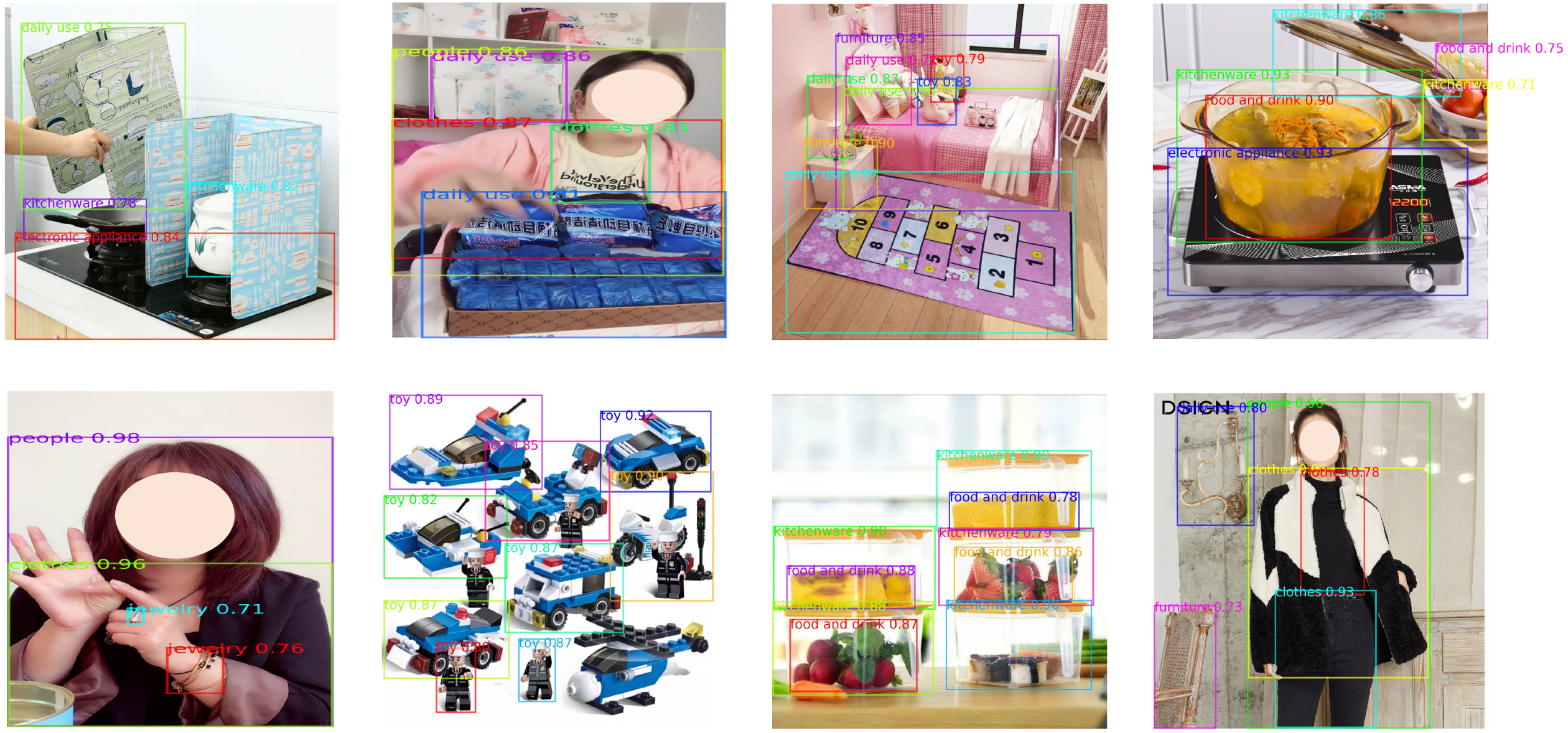}
  \caption{Qualitative examples of object detection.}
  \label{fig:supp_fig_det}
  \vspace{-0.17in}
\end{figure*}

\subsubsection{Object Detection}
On object detection, an image usually contains multiple foreground instances. Hence, we increase the query number to $40$. The position and type embeddings of newly added $20$ queries are copied directly from the pre-trained ones. Besides, the box prediction head is implemented as a 3-layer MLP as in DETR~\cite{carion2020end}. For the compared baselines, we initialize the image encoder with the pre-trained weight and train other parameters from scratch. The collected object detection dataset contains 146,813 samples from 18 classes for training and 25,909 samples for evaluation. During finetuning, the image resolution is increased to $640 \times 640$ and the batch size is set to 128 on 8 A100 GPUs. We finetune the entire model for 80 epochs, with the learning rate of $1e{-5}$ for the image encoder and $8e^{-5}$ for the remaining ones.

\subsection{Additional Experimental Results}
\subsubsection{Zero-Shot Image-Conditioned Grounding}
\label{sec:supp_image_ground}
The image-conditioned Grounding requires the model to localize the instance depicted by a query image. Different traditional visual grounding, it is suitable when the query is difficult to describe through text. Since ECLIP also incorporates image prompts during pretraining, we can thus transfer it to image-conditioned grounding without further finetuning. To this end, we construct a grounding dataset consisting of 100K samples, where each one has a test and query image. The test image contains multiple instances and the query image has only one instance that needs to be grounded. During evaluation, each query is first embedded into $g_I(v_{\text{cls}})$ by the image encoder. Similar to Section~\ref{sec:ground}, we can estimate the 2D similarity map and then rank each proposal to select the candidate with maximum $r(\cdot)$. Table~\ref{tab:supp_grounding} shows detailed results for ECLIP and compared baselines. Notably, ECLIP is still highly competitive in image-conditioned visual grounding. By contrast, approaches that learn global representation ignore fine-grained feature alignment and thus achieve worse performance.

\subsubsection{Ablation of Instance Query Number}
We also explore the effect of instance query number $T$ of the ECLIP decoder. Due to the huge training costs, all experiments are conducted on a smaller pretraining subset that includes only 5M images. As shown in Table~\ref{tab:supp_abla}, increasing the query number from $20$ to $40$ will slightly boost performance on instance-level product retrieval. It is intuitive because more queries bring the potential to focus on more instances. Accordingly, the computation and GPU memory load will increase as well. Therefore, we set $T=20$ during pretraining for all other experiments.

\subsubsection{The effect of Slot-Attention} 
The slot-attention distributes each visual token to one of $T$ queries according to their similarity, which explicitly divides an image into $T$ different parts. Such a mechanism will encourage the positive query to focus on the region that contains correlated instances and the negative ones to be distracted by the background. In contrast, the cross-attention weights tend to be smoothed over all image tokens (See Figure~\ref{fig:supp_fig_ground}). We verify the effectiveness of slot-attention via replacing it with cross-attention in the decoder layer. The results in Table~\ref{tab:supp_ablation2} present clear performance gain brought by slot-attention. We also explore the effect of loss terms $\mathcal{L}_{itm}$ and $\mathcal{L}_{\mathcal{R}}$. As in Table~\ref{tab:supp_ablation2}, all the components contribute to the final performance.

\section{More Visualization Results}
\label{sec:supp_vis}
In order to qualitatively demonstrate the strong generalization of ECLIP on downstream E-commerce tasks, we provide more visualizations in this section.

\vspace{0.03in}

\noindent \textbf{Object Detection}. The finetuned object detection results achieved by ECLIP are shown in Figure~\ref{fig:supp_fig_det}. It further illustrates the promising fine-grained understanding capability in real-world E-commerce applications. Even for some complex scenes, ECLIP is able to successfully detect the various product instances appearing in the given image.

\begin{table}[t]
    \centering
    \resizebox{\linewidth}{!}{
    \begin{tabular}{c|cccc}
    \toprule
    $T$ & Peak GPU Memory  & mAP@10 &  mAP@50 &  mAP@100 \\
    \midrule
    10 &  47.1 GB  &  87.5  &  82.7 & 81.3 \\
    20 &  48.6 GB  &  89.6  & 84.6  & \textbf{82.2}  \\
    40 & 52.3 GB  & \textbf{89.9}  & \textbf{84.7}   & 81.9  \\
    \bottomrule
    \end{tabular}}
    \vspace{-0.5em}
    \caption{Ablation results of different number of instance queries on instance-level product retrieval (Visual Modality).}
    \label{tab:supp_abla}
    \vspace{-0.5em}
\end{table}

\begin{table}[t]
    \centering
    \resizebox{\linewidth}{!}{
    \begin{tabular}{lcccc}
    \toprule
    \multirow{2}{*}{Setting} & \multicolumn{2}{c}{Visual Grounding} & \multicolumn{2}{c}{Coarse-Level Retrieval} \\
    \cmidrule(lr){2-3} \cmidrule(lr){4-5} & Acc@0.5 & Acc@0.7   & mAP@1 & mAP@5  \\
    \midrule
    Cross-Attention &  73.9  &  66.3 &  53.2  & 55.4    \\
    Slot-Attention & \textbf{78.7}    & \textbf{70.5}   & \textbf{54.7}   & \textbf{57.8}  \\
    \midrule
    Base &  77.4 &  69.6  &   53.2 &    56.4 \\
    Base+$\mathcal{L}_{itm}$ & 78.1  &  70.0  &  54.2  &  56.9   \\
    Base+$\mathcal{L}_{itm}$+$\mathcal{L}_{\mathcal{R}}$ & \textbf{78.7}  &   \textbf{70.5} & \textbf{54.7}   &  \textbf{57.8}   \\
    \bottomrule
    \end{tabular}
    }
    \vspace{-0.5em}
    \caption{\small Ablation results of different loss terms and slot attention.}
    \label{tab:supp_ablation2}
    \vspace{-1.2em}
\end{table}

\begin{figure*}[t]
  \centering
  \includegraphics[width=0.88\linewidth]{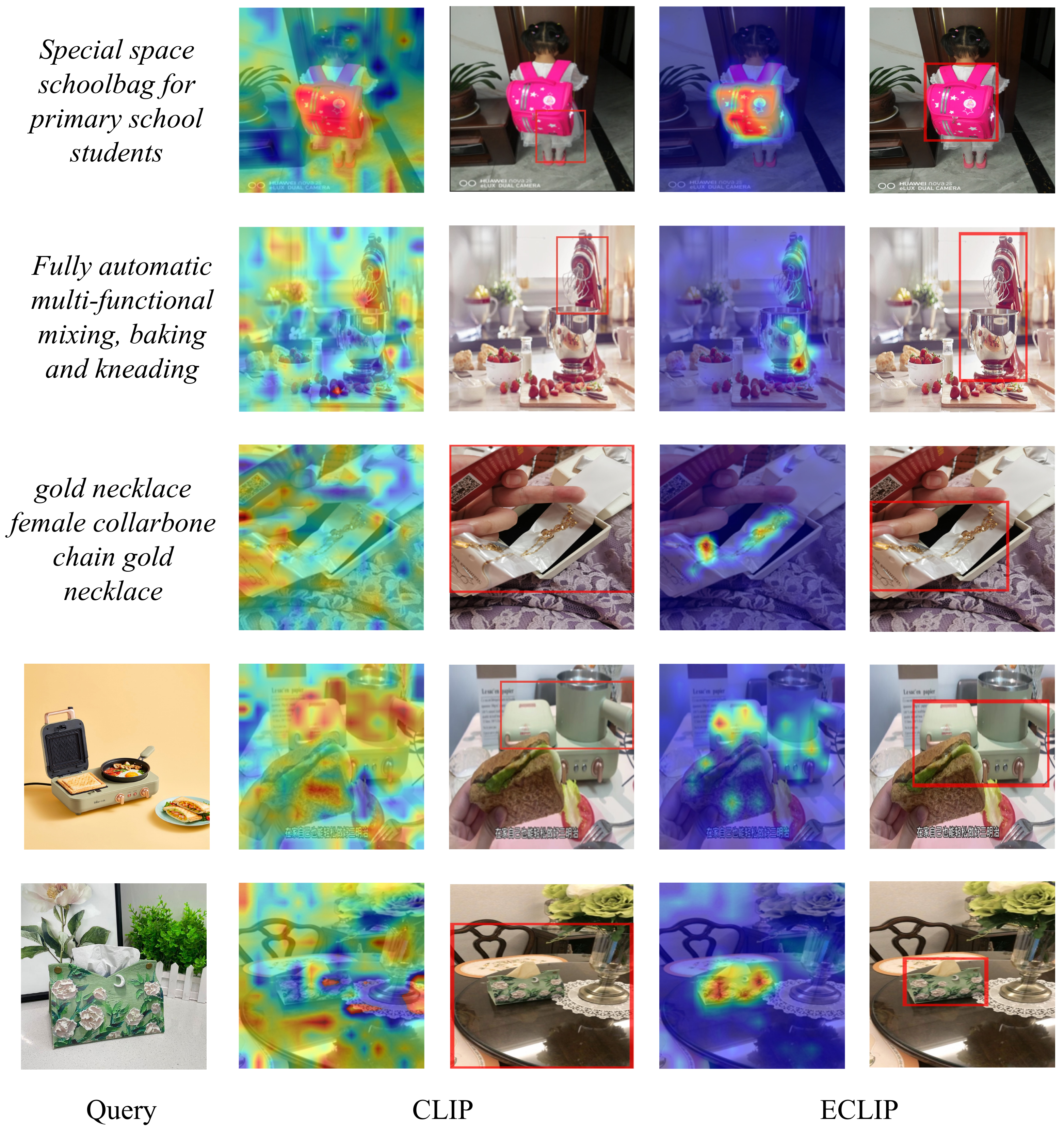}
  \caption{Visualizations of zero-shot text- and image-conditioned visual grounding achieved by CLIP and ECLIP.}
  \label{fig:supp_fig_ground}
  \vspace{-1.5em}
\end{figure*}

\begin{figure*}[t]
  \centering
  \includegraphics[width=0.88\linewidth]{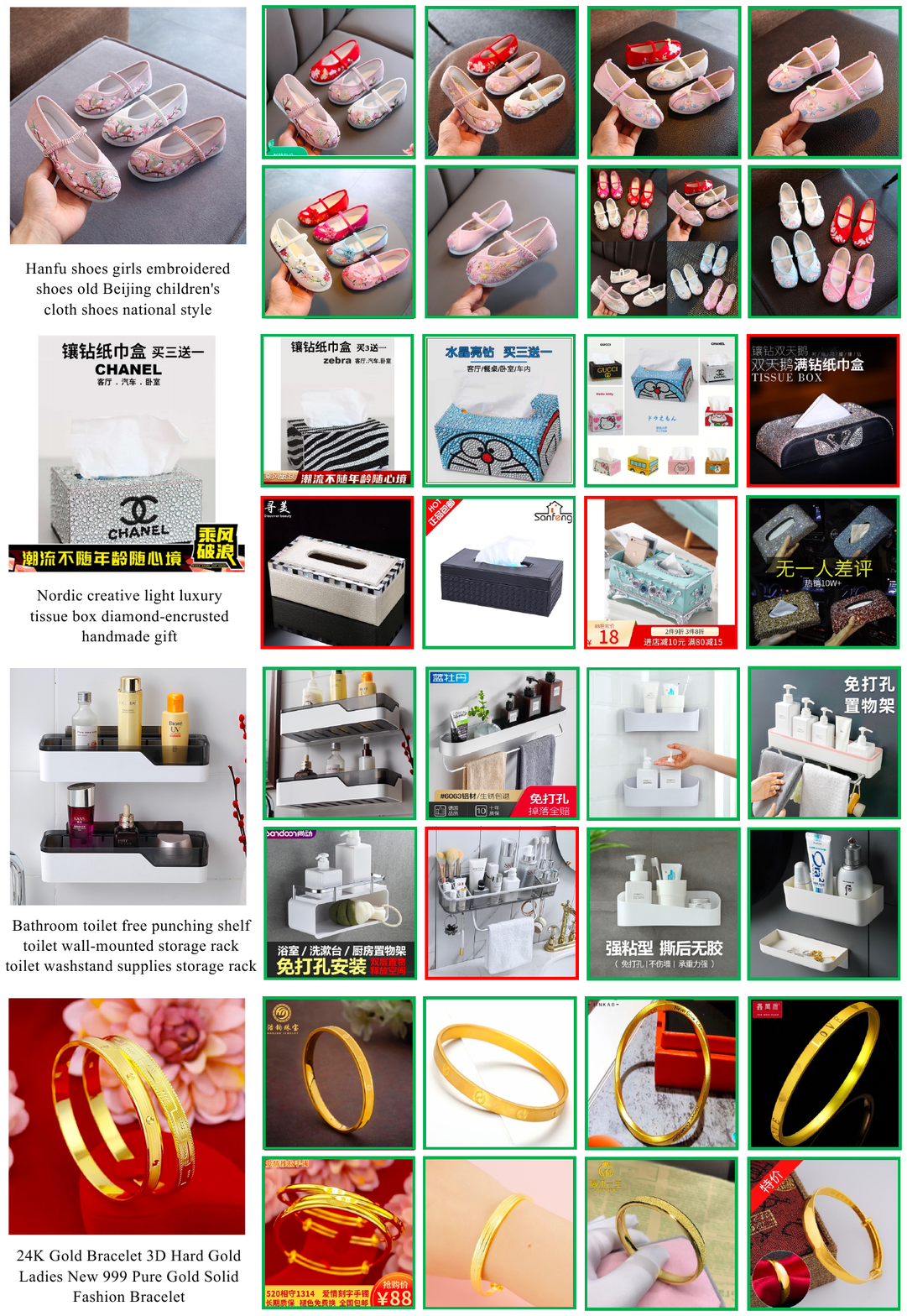}
  \caption{Visualizations of zero-shot coarse-level retrieval results achieved by ECLIP.}
  \label{fig:supp_fig_coarse}
\end{figure*}

\begin{figure*}[t]
  \centering
  \includegraphics[width=0.86\linewidth]{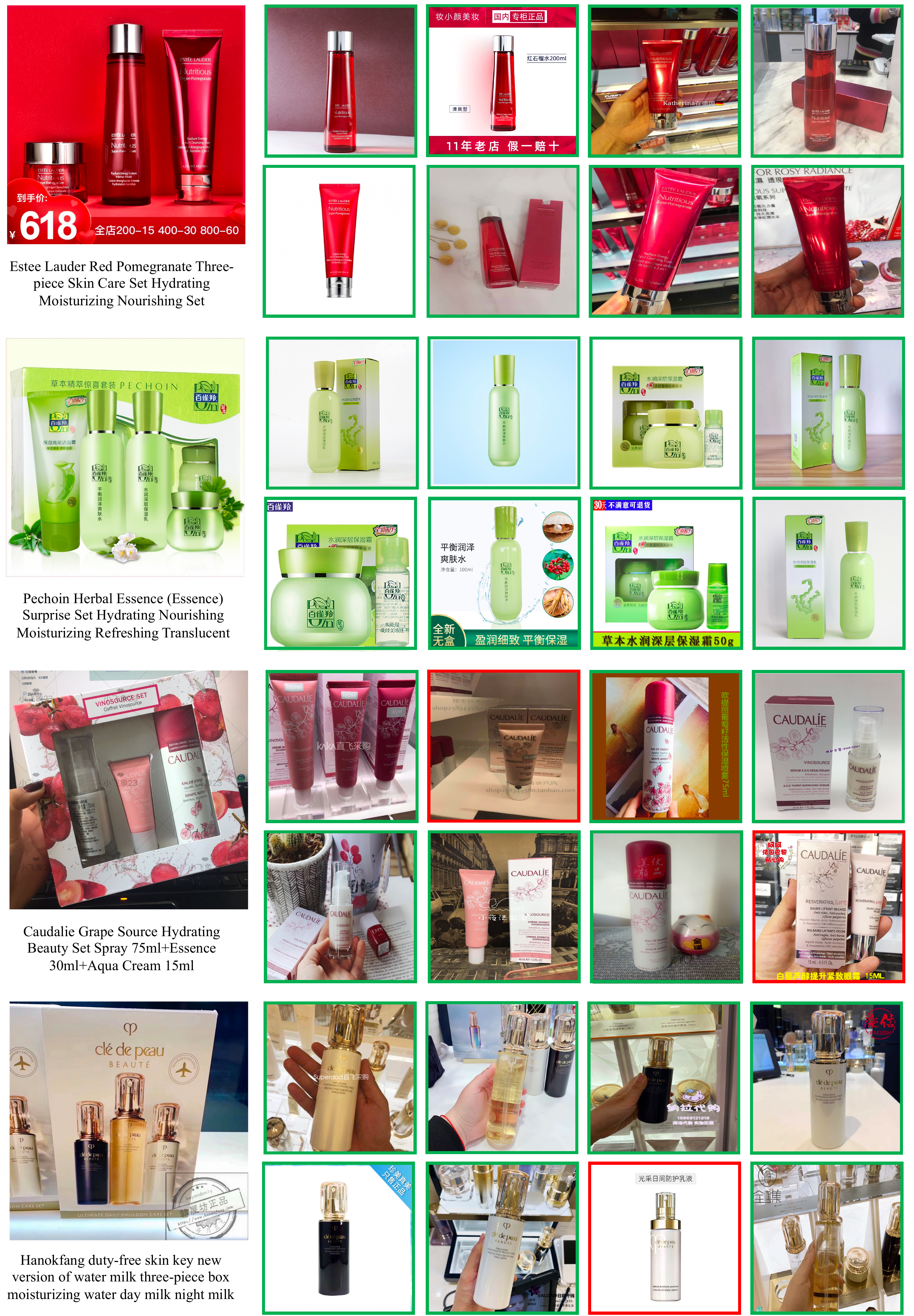}
  \caption{Visualizations of zero-shot instance-level retrieval results achieved by ECLIP.}
  \label{fig:supp_fig_ins}
\end{figure*}

\vspace{0.03in}

\noindent \textbf{Zero-Shot Visual Grounding}. We here provide additional qualitative examples of zero-shot visual grounding. In Figure~\ref{fig:supp_fig_ground}, we illustrate the comparison of text- and image-conditioned grounding with CLIP. Obviously, ECLIP can properly attend to the desired instance depicted by the text or image query. Moreover, compared to CLIP, the similarity map is highly concentrated on the target regions of interest. However, CLIP's similarity score map is distracted by many unrelated background instances, due to ignorance of the instance-level modeling. 

\vspace{0.03in}


\noindent \textbf{Zero-Shot Product retrieval}. We also illustrate the coarse- and instance-level product retrieval examples in Figure~\ref{fig:supp_fig_coarse} and Figure~\ref{fig:supp_fig_ins}. The images marked in green are the samples correctly retrieved, while images marked in red are mismatched ones. It can be observed that ECLIP successfully returns satisfactory retrieval results. Especially for challenging instance-level retrieval, it can still recall the existing product instances in a query image from a large gallery set.

\vspace{-0.6em}

\section{Boarder Impact}
\label{sec:supp_board}
This work provides a novel perspective for learning prompt-based visual representation. The unique contributions of ECLIP as follows: (1) This work mainly focuses on E-commerce scenarios. E-commerce oriented model, despite its high practical importance, still remain inadequately studied. We identify the unique challenge by comparing natural / product images, \emph{i.e.}, the gap between the demand for instance-level representation and the lack of box annotations in large-scale raw E-commerce data (See Figure~\ref{fig:fig1}). (2) The proposed instance decoder innovatively correlates the multi-modal prompt with input queries and adopts the slot-attention to implicitly force each query to attend to a specific image region. The developed two proxy tasks can fully exploit the natural characteristics of E-commerce data itself as supervision. These novel designs collectively enable ECLIP to effectively ground a desired instance (see Figure~\ref{fig:supp_fig_ground}). In contrast, the existing VLP models (e.g., X-VLM~\cite{zeng2021multi}, GLIP~\cite{Li_2022_CVPR}, MDETR~\cite{kamath2021mdetr}, etc) obtain such ability by relying on object-level annotations.

\end{document}